# Rotation, Scaling and Translation Analysis of Biometric Signature Templates

Aman Chadha, Divya Jyoti, M. Mani Roja
*Thadomal Shahani Engineering College, Mumbai, India*
aman.x64@gmail.com

**Abstract**

*Biometric authentication systems that make use of signature verification methods often render optimum performance only under limited and restricted conditions. Such methods utilize several training samples so as to achieve high accuracy. Moreover, several constraints are imposed on the end-user so that the system may work optimally, and as expected. For example, the user is made to sign within a small box, in order to limit their signature to a predefined set of dimensions, thus eliminating scaling. Moreover, the angular rotation with respect to the referenced signature that will be inadvertently introduced as human error, hampers performance of biometric signature verification systems. To eliminate this, traditionally, a user is asked to sign exactly on top of a reference line. In this paper, we propose a robust system that optimizes the signature obtained from the user for a large range of variation in Rotation-Scaling-Translation (RST) and resolves these error parameters in the user signature according to the reference signature stored in the database.*

**Keywords:** *rotation; scaling; translation; RST; image registration; signature verification.*

## 1. Introduction

The aim of a biometric verification system is to determine if a person is who he/she purports to be, based on one or more intrinsic, physical or behavioral attributes. This trait or biometric attribute can be the signature, voice, iris, face, fingerprint, hand geometry etc.

A simple biometric system has a sensor module, a feature extraction module, a matching module and a decision making module. The sensor module acquires the biometric data of an individual. In this case, the digital pen tablet functions as the sensor. In the feature extraction module, the acquired biometric data is processed to extract a feature set that represents the data. For example, the position and orientation of certain specific points in a signature image are extracted in the feature extraction module of a signature authentication system. In the matching module, the extracted feature set is compared against that of the template by generating a matching score. In this module, the number of matching points between the acquired and reference signatures are determined, and a matching score is obtained. Decision-making involves either verification or identification. In the decision-making module, the user's claimed identity is either accepted or rejected based on the matching score, i.e., verification. Alternately, the system may identify a user based on the matching scores, i.e., identification [1],[11].

Signature recognition is one of the oldest biometric authentication methods, with wide-spread legal acceptance. Handwritten signatures are commonly used to approbate the contents of a document or to authenticate a financial transaction [1]. A trivial method of signature verification is visual inspection. A manual comparison of the two signatures is done and the given signature is accepted if it is sufficiently similar to the reference signature, for example, on a credit-card. In most scenarios, where a signature is used as the means of authentication, no verification takes place at all due to the entire process being excessively time intensive and demanding. An automated signature verification process will help improve the current situation and thus, eliminate fraud. Well-known biometric methods include iris, retina, face and fingerprint based identification and verification. Even though human features such as iris, retina and fingerprints do not change over time and have low intra-class variation, i.e., the variations in the respective biometric attribute are low, special and relatively expensive hardware is needed for data acquisition in such systems. An important advantage of signatures as the human trait for biometric authentication over other attributes is their long standing tradition in many commonly encountered verification tasks. In other words, signature verification is already accepted by the general public. In addition, it is also relatively less expensive than the other biometric methods [1],[2].

The difficulties associated with signature verification systems due to the extensive intra-class variations, make signature verification a difficult pattern recognition problem. Examples of the various alterations observed in the signature of an individual have been illustrated in Fig. 1.

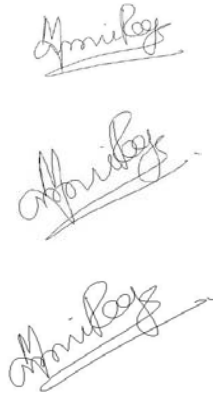

**Figure 1: Intra-class variations, i.e. variations in the signature of an individual**

Depending on the data acquisition method, automatic signature verification can be divided into two main types: off-line and on-line signature verification. The most accurate systems almost always take advantage of dynamic features like acceleration, velocity and the difference between up and down strokes [3]. This class of solutions is called on-line signature verification. However in the most common real-world scenarios, because such systems require the observation and recording off the signing process, this information is not readily available. This is the main reason, why static signature analysis is still in focus of many researchers. On-line signature verification uses special hardware, such as a digitizing tablet or a pressure sensitive pen, to record the pen movements during writing. In addition to shape, the dynamics of writing are also captured in on-line signatures, which is not present in the 2-D representation of the signature and hence it is difficult to forge. Off-line methods do not require special acquisition hardware, just a pen and a paper, and are therefore less invasive and more user friendly. In the past decade a bunch of solutions has been introduced, to overcome the limitations of off-line signature verification and to compensate for the loss of accuracy [2],[3]. In off-line signature verification, the signature is available on a document which is scanned to obtain its digital image. In all applications where handwritten signatures currently serve as means of authentication, automatic signature verification can be used such as cashing a check, signing a credit card transaction or authenticating a legal document. Basically, any system that uses a password can instead use an on-line signature for access. The advantages are such systems are obvious – a signature is more difficult to steal or guess than a password and is also easier to remember for the user.

However, the high level of intra-class variations in signatures, as shown in Fig. 1, hinder the performance of signature verification systems and thus minimize the accuracy of such systems. Hence, to reduce errors and the inefficiency problems associated with these systems, the intra-class variations in the signatures need to be minimized. This involves eliminating or reducing the rotation, scaling and translation factors between the reference and the test signature images. Fig. 2 shows the diagram of a typical signature verification system with rotation, scaling and translation (RST) cancellation. The reference image within the database and the user image act as inputs to the system. Feature extraction is done from the reference signature which describes certain characteristics of the signature and stored as a template. For verification, the same features are extracted from the test signature and compared to the template.

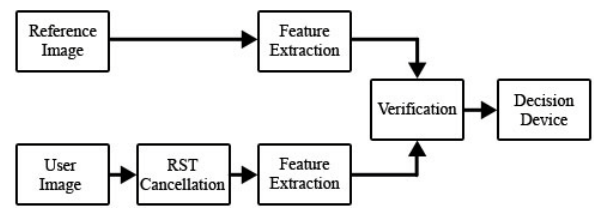

**Figure 2: A typical signature verification system with RST cancellation**

It should be noted that a distinct advantage of the proposed system, illustrated in Fig. 2, is that it does not require multiple signature reference samples for training in order to achieve high levels of accuracy. Previous work by researchers has witnessed the use of affine transformation for calculating the angular rotation between two images, the scaling and translation [4]-[7].

In this paper, we propose the use of the concept of correlation to identify the rotation and a simple cropping method to eliminate scaling and translation, thereby creating an optimum template after subjecting the user image to RST correction.

## 2. Idea of the proposed solution

The foremost concern is fetching the angle of rotation between the user and the reference images. In order to achieve this, the concept of correlation is deployed. The term "correlation" is a statistical measure, which refers to a process for establishing whether or not relationships exist between two variables [8]. The maximum value of cross-correlation between the original, i.e., the reference image and the user image is found by means of repetitive iterations involving the calculation of the cross-correlation between the two images in question.

The proposed algorithm essentially finds the cross-correlation between original image and the user image. If $X(m, n)$ is reference image and $Y(m, n)$ is the user image then the cross-correlation *r* between X and Y is given by the following equation:

$$r = \sum_m \sum_n (X_{mn} - X_0)(Y_{mn} - Y_0) \quad (1)$$

Minimum value of *r* indicates dissimilarity of images and for the same image (autocorrelation) it will have a peak value so as to indicate 'maximum correlation'. $X_0$ and $Y_0$ represent mean of Image X and Y respectively.

We use normalized cross-correlation to simplify analysis and comparisons of coefficient values corresponding to the respective angular values. Min-max normalization is the procedure used to obtain normalized cross-correlation [9]. Min-max normalization preserves the relationships among the original data values. The normalization operation transforms the data into a new range, generally [0, 1]. Given a data set $x_i$, such that i = 1, 2, . . . , n, the normalized value *x'* is given by the following equation:

$$x' = \frac{x - \min(x_i)}{\max(x_i) - \min(x_i)} \quad (2)$$

The second aim is to deal with the translation associated with the images. This is achieved by a simple cropping technique. Initially, we calculate the number of rows and columns bordering the signature pixels within the image. The image devoid of these rows and columns is extracted. The result is an image consisting of only the signature pixels. Additional background surrounding the image is thus eliminated.

Third factor is the scaling between the two images. For calculation of the scaling factor, the cropped images obtained during translation are utilized. The size of the reference image divided by the size of the user image gives the scaling ratio.

The proposed solution is illustrated by Fig. 3.

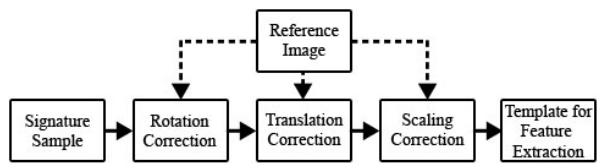

**Figure 3: Schematic block diagram of the proposed system**

## 3. Implementation steps

### 3.1. Image acquisition and pre-processing

Our image acquisition is inherently simple and does not employ any special illumination. The system implemented here uses a digital pen tablet, namely, WACOM Bamboo [10], as the data-capturing device. The pen has a touch sensitive switch in its tip such that only pen-down samples (i.e., when the pen touches the paper) are recorded. The database consists of a set of signature samples of 90 people. For each person, there are 9 test images and 1 training or reference image in the database. Upon signature acquisition, the next step is colour normalization and binarization. Colour normalization is the conversion of the image from the RGB form to the corresponding Grayscale image. Binarization is the conversion of this grayscale image to an image consisting of two luminance elements, namely, black and white. On completion of the image acquisition and pre-processing stage, the resultant image thus obtained, becomes ready for the corrective phases: rotation, scaling and translation cancellation.

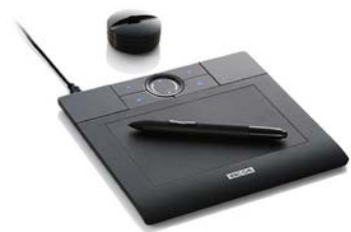

**Figure 4: Wacom Bamboo Digital Pen Tablet**

### 3.2. Rotation correction

While collecting signature samples, it was observed that users gave consecutive samples having angular variations approximately from –60° to +60°. Hence, before feature extraction, the user image should be aligned with the reference image. For simplicity in computing rotation angle, we choose to align the reference image with the trial image fetched from the user, i.e., the user image.

The preprocessed reference image is cropped in order to extract only the signature pixels without any additional background and used for all further computations. In order to make the program time efficient and less resource intensive, two stages of rotation correction are applied. The first stage is designed to offer a relatively lower resolution of 5° so as to offer an approximate value of the angle of rotation. In contrast, the second stage is designed for a comparatively higher resolution. Within a range of +3° to –3° of the approximate value, a resolution of 1° is selected for a more precise value of the rotation angle.

After pre-processing, the user image is then rotated by 5° within the range of –60° to +60° in successive iterations. Cross-correlation values between the reference image and the user image are recorded on completion of each iteration of the rotation process. The maximum cross-correlation value refers to the correct angle of rotation within a 5° range, further, after the approximate angle value is obtained, +3° or –3° of this angle can be inspected for maximum correlation value which corresponds to angle of rotation accurate to up to 1°. The user image is rotated by the negative of the angle thus obtained, and then subjected to feature extraction. Thus, rotation cancellation is achieved.

The steps involved in the rotation correction process can be summarized as follows:

1) *Obtain user image and the reference image.*
2) *Carry out pre-processing by converting both images to grayscale and performing normalization.*
3) *Trim the reference signature to remove any excess background; this will act as the template.*
4) *Starting with the angle as –60°, in increments of 5°, record normalized correlation values between pre-processed reference image and user image.*
5) *If angle is less than or equal to 60°, go to step 4.*
6) *Maximum correlation value corresponds to angle within a 5° range. Let this angular value be x°.*
7) *Starting with the angle as (x – 3)°, in increments of 1°, record normalized correlation values between the preprocessed reference image and the user image.*
8) *If angle is less than or equal to (x + 3)°, go to step 7.*
9) *Correct the user image by the obtained angle and proceed for further correction, if required.*

Fig. 5 shows a reference image and the corresponding image rotated by 20.9°.

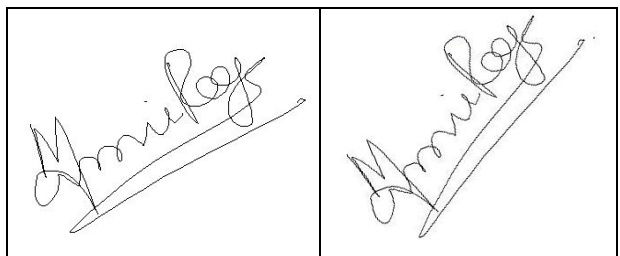

**Figure 5: Reference image and the correponding rotated image**

### 3.3. Scaling correction

An end user often modifies his/her signature according to the size of signing box. For smaller spaces, the signature may be compressed, for no space limitation, the sign may be enlarged. Thus, before extraction of feature points, it is essential that any scaling, if present in the test sample, be removed. Upon trimming both images, the ratio of height gives Y scaling and ratio of width gives X scaling. However, to resize the user image and make it the same size as the registered image, either of the scaling ratios can be used. For a rotation range of –60° to +60°, height was observed to vary significantly as compared to the length. Hence, Y scaling was chosen as the scaling ratio. To account for scaling, the above mentioned cropping technique is applied to both the user as well as reference image. Scaling ratio is calculated by the following equation:

$$\text{Scaling ratio} = \frac{\text{Size of the reference image}}{\text{Size of the test image}} \quad (3)$$

The user image is resized as per the obtained scaling ratio and then sent to the feature extraction segment. Fig. 6 shows a reference image and the corresponding image down scaled by a scaling ratio of 1.4045.

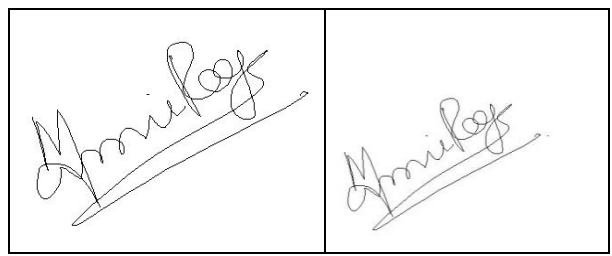

**Figure 6: Reference image and the corresponding down-scaled image**

### 3.4. Translation correction

On the apparatus used for taking signature input, the user is free to sign without using any fixed starting point. This may introduce translation in X and/or Y direction, having a maximum value equal to the width or height of the signature canvas respectively. The boundary conditions for translation error are computed assuming that the user starts to sign from the edge.

This problem is overcome by cropping the pre-processed reference image so as to extract only the signature pixels without any additional background. This cropping process truncates the extra background region by trimming the image canvas. Thus, translation is removed completely. For representational purposes, bottom left corner of test image is assumed to be origin.

The number of columns from left and number of rows from the bottom, which contain no black pixels corresponding to the actual signature, i.e., which consist solely of image-background, are counted. These values give X translation and Y translation respectively. Fig. 7 shows a reference image and the corresponding image translated by 35px along X-Axis and 9px along Y-Axis.

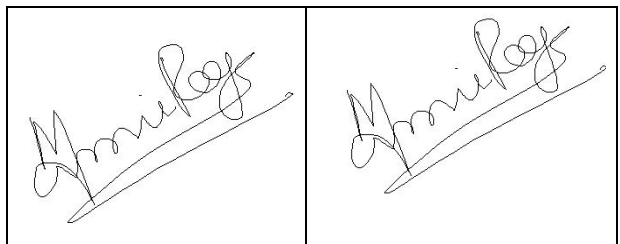

**Figure 7: Reference image and the corresponding translated image**

### 3.5. Combined Rotation–Scaling–Translation

It is easy to manipulate the samples to get pure rotation, translation and scaling, however, for actual signatures, all the above mentioned factors are altered simultaneously. Hence, rotation, translation and scaling corrections are applied in the same order.

Rotation correction precedes translation correction as the assumed origin at bottom left corner also gets rotated and translation effects cannot be eliminated unless the origin is returned to bottom left as accurately as possible. Therefore, rotation correction needs to be performed first as the scaling ratio calculated by the pure scaling method is not consistent with scaling ratio of the rotated image, as shown in Fig. 8.

Consequently, the effectiveness of scaling correction depends, to a large extent, on the percentage error obtained during rotation correction.

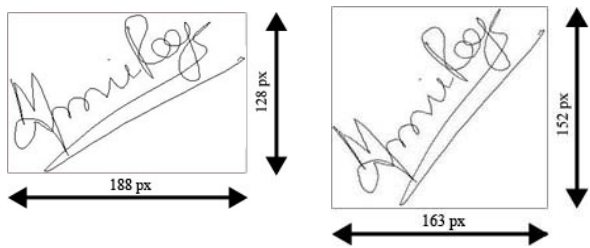

**Figure 8: Change in the width and height of the image before and after rotation correction**

After correcting the angle of rotation, the user image pre-processed copy is cropped to eliminate translation and the image so obtained is a case of pure scaling which has been discussed above.

Thus, rotation–scaling–translation cancellation is achieved.

## 4. Results

### 4.1. Rotation

Results obtained for pure rotation have been tabulated as follows:

**Table 1: Results obtained for pure rotation**

| Signature Samples | Actual Angle | Detected Angle | % Error |
|---|---|---|---|
| Sample 1 | -60 | -60 | 0 |
| Sample 2 | -48 | -48 | 0 |
| Sample 3 | -20 | -20 | 0 |
| Sample 4 | -6 | -6 | 0 |
| Sample 5 | 0 | 0 | 0 |
| Sample 6 | 4 | 4 | 0 |
| Sample 7 | 13 | 13 | 0 |
| Sample 8 | 27 | 27 | 0 |
| Sample 9 | 37 | 37 | 0 |
| Sample 10 | 59 | 60 | 1.69 |

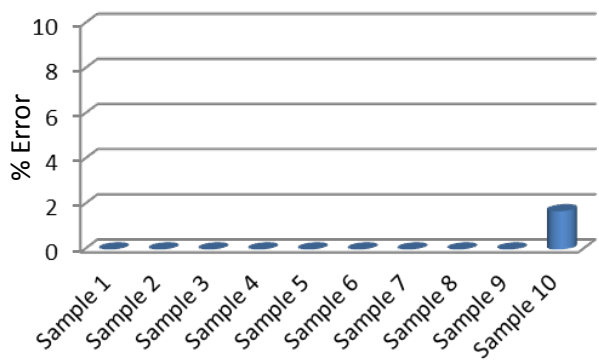

**Figure 9: Plot of % error values in case of pure rotation for various samples**

### 4.2. Scaling

Results obtained for pure scaling have been tabulated as follows:

**Table 2: Results obtained for pure scaling**

| Signature Samples | Actual Scaling Ratio | Detected Scaling Ratio | % Error |
|---|---|---|---|
| Sample 1 | 7.69 | 10.55 | 37.2 |
| Sample 2 | 5 | 5.70 | 14 |
| Sample 3 | 4 | 4.22 | 5.5 |
| Sample 4 | 2.17 | 2.27 | 4.6 |

| Signature Samples | Actual Scaling Ratio | Detected Scaling Ratio | % Error |
|---|---|---|---|
| Sample 5 | 1.28 | 1.34 | 4.7 |
| Sample 6 | 1 | 1.05 | 5 |
| Sample 7 | 0.63 | 0.66 | 4.8 |
| Sample 8 | 0.54 | 0.57 | 5.6 |
| Sample 9 | 0.48 | 0.49 | 2.1 |
| Sample 10 | 0.31 | 0.33 | 6.5 |

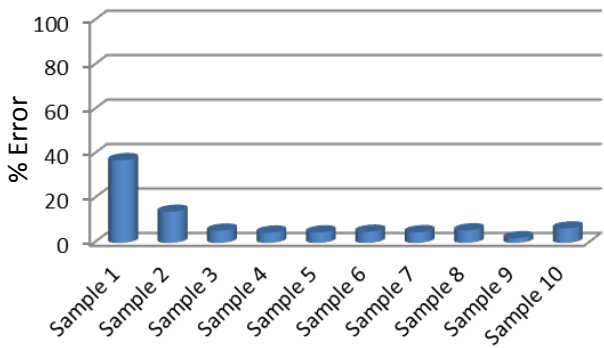

**Figure 10: Plot of % error values in case of pure scaling for various samples**

### 4.3. Translation

Results obtained for pure translation have been tabulated as follows:

**Table 3: Results obtained for pure translation**

| Signature Samples | Actual Translation | Recovered Translation | % Error |
|---|---|---|---|
| Sample 1 | 0,5 | 0,5 | 0 |
| Sample 2 | 5,5 | 5,5 | 0 |
| Sample 3 | 10,0 | 10,0 | 0 |
| Sample 4 | 15,10 | 15,10 | 0 |
| Sample 5 | 0,25 | 0,25 | 0 |
| Sample 6 | 25,25 | 25,25 | 0 |
| Sample 7 | 25,50 | 25,50 | 0 |
| Sample 8 | 50,50 | 50,50 | 0 |
| Sample 9 | 50,100 | 50,100 | 0 |
| Sample 10 | 150,150 | 150,150 | 0 |

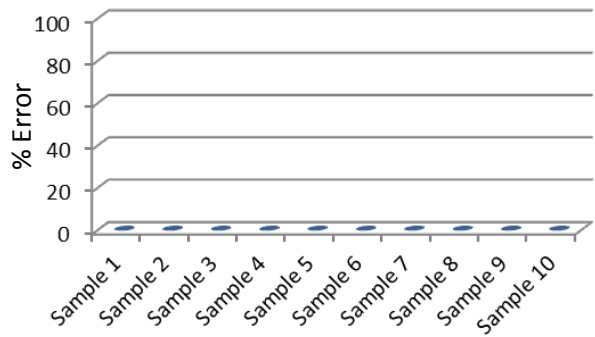

**Figure 11: Plot of % error values in case of pure translation**

### 4.4. Rotation-Scaling-Translation

Results obtained upon combining rotation, scaling and translation have been tabulated as follows:

**Table 4: Results obtained on combining rotation, scaling and translation**

| Signature Samples | Actual Parameters | | Detected Parameters | |
|---|---|---|---|---|
| | *Rotation* | *Scaling* | *Rotation* | *Scaling* |
| Sample 1 | 50 | 1.67 | 52 | 1.90 |
| Sample 2 | 12 | 1.33 | 10 | 1.39 |
| Sample 3 | 31 | 1.11 | 34 | 1.25 |
| Sample 4 | -40 | 0.91 | -42 | 0.94 |
| Sample 5 | -30 | 0.8 | -32 | 0.82 |

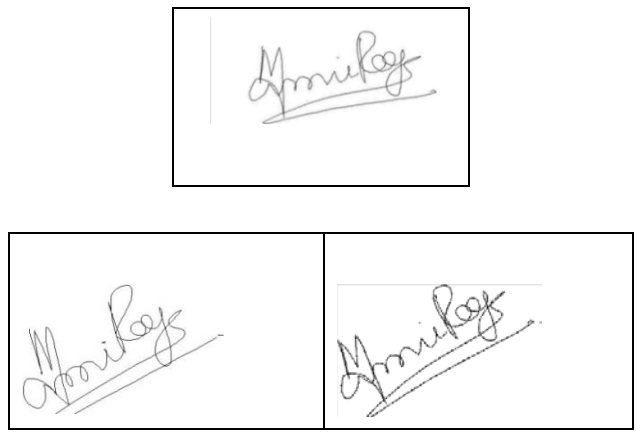

**Figure 12: User image before RST correction (top); reference image (bottom-left) and the user image after RST Correction (bottom-right)**

## 5. Conclusion

The system has been designed to correct variations in angle of rotation, in the range of –60° to +60°. However, it can be extended to cover the entire 360° planar rotation, to design a fool proof system capable of creating an optimum template after RST cancellation, even in a situation where the input pad may have been inverted. On similar lines, the resolution of rotation can be improved from 1° to 0.5° or 0.25° or even more, with the trade-off being increased program execution times.

Pure scaling and pure translation can be detected accurately as long as signature pixels do not go beyond the defining boundaries of the template. For the signatures used, a maximum translation of 200 pixels was detected along X and Y axes. Maximum scaling ratio was found to be 0.55. However, maximum variance of both translation and scaling may show slight variations from one signature to another.

For combined RST, it was experimentally observed that the correlation approach tends to be less reliable with significant increase or decrease in the scaling ratio. Signature images used for testing gave optimum result for scaling ratio, i.e., within 0.67 to 1.33, however, the scaling range giving angle and translation accurately may increase or decrease depending on the signature sample under test.

Thus, an optimum template was generated by the proposed system after subjecting the user image to RST correction with respect to the reference image.